\documentclass[runningheads]{llncs}

\usepackage[english]{babel}

\usepackage{graphicx}
\usepackage{xcolor}
\usepackage{subcaption}
\usepackage{amsmath}
\usepackage{amsfonts}
\usepackage{hyperref}

\graphicspath{{build/images/}}

\newcommand{\figref}[1]{Fig.~\ref{#1}}

\let\oldeqref\eqref
\renewcommand{\eqref}[1]{eq.~\oldeqref{#1}}

\DeclareMathOperator*{\argmax}{arg\,max}

\begin{document}

\title{Optimizing Revenue Maximization and Demand Learning in Airline Revenue Management}

\titlerunning{Optimizing revenue and demand learning in airline RM}

\author{Giovanni Gatti Pinheiro\inst{1, 2} \and
Michael Defoin-Platel\inst{1} \and
Jean-Charles Regin\inst{2}}

\authorrunning{G. Gatti Pinheiro et al.}

\institute{Amadeus SAS, 821 Avenue Jack Kilby, 06270 Villeneuve-Loubet, France \and Universite de la Cote d'Azur}
\maketitle              
\begin{abstract}
Correctly estimating how demand respond to prices is fundamental for airlines willing to optimize their pricing policy. Under some conditions, these policies, while aiming at maximizing short term revenue, can present too little price variation which may decrease the overall quality of future demand forecasting. This problem, known as earning while learning problem, is not exclusive to airlines, and it has been investigated by academia and industry in recent years. One of the most promising methods presented in literature combines the revenue maximization and the demand model quality into one single objective function. This method has shown great success in simulation studies and real life benchmarks. Nevertheless, this work needs to be adapted to certain constraints that arise in the airline revenue management (RM), such as the need to control the prices of several active flights of a leg simultaneously. In this paper, we adjust this method to airline RM while assuming unconstrained capacity. Then, we show that our new algorithm efficiently performs price experimentation in order to generate more revenue over long horizons than classical methods that seek to maximize revenue only.

\end{abstract}

\section{Introduction}
When optimizing the pricing policy, modern revenue management systems consider only the revenue-maximizing objective, ignoring the long-term effects on the future learning of the demand behavior. For example, in certain cases, the optimized pricing policy may have little price variation making it difficult to correctly estimate the customer's price sensitivity, impacting the quality of future forecasts. The proper estimation of the customer's price sensitivity is fundamental to airlines, in fact, researchers have found that a bias of $\pm 20\%$ in the estimation of the price sensitivity can reduce revenue by up to $4\%$ \cite{fiig2019can}.

To learn the customer's price sensitivity, the revenue management system (RMS) needs to charge different prices that may not be optimal with respect to revenue maximization (i.e., price experimentation), compromising short term revenue with the hope that the information gained about the demand behavior will allow the RMS to collect more revenue in the long run. Thus, we have two competing objectives and a trade-off must be found. This is known as the \textit{earning while learning} (EWL) problem. 

The EWL problem is not exclusive to the airline industry. It applies to any seller that needs to price its products in face of an unknown demand behavior. This is one of the reasons why this problem attracted the attention from academia and industry in recent years. In one of the most recent and promising methods, the authors propose to unify revenue maximization and demand learning into one single objective function \cite{elreedy2021novel}. They find that their method outperforms several others in literature over simulation studies and different real-world benchmarks. However, their method is limited to the optimization of a single product (i.e., a flight), which makes it not directly exploitable by airlines.

We design an adaptation of their method to the airline revenue management (RM) environment, which must control the pricing policy of several active flights simultaneously. We focus on specific characteristics of airline RM, such as the aspect that the historical booking data used for demand model learning is collected by many parallel flights, and thus, the optimization of the new objective must consider the effects across different flights. Moreover, for the sake of simplicity, we assume the price optimization problem under unconstrained capacity (which simplifies significantly the price optimization process), and that a simplified demand model is estimated (which facilitates interpretation). Applying our method to real-world systems is beyond the scope of this work, and more research is encouraged.

Our work is organized as follows. In Section \ref{background}, we present the simplified single leg problem and how RMS typically estimates the demand behavior and optimizes the pricing policy. Then, in Section \ref{related-work}, we present some related work on the investigation of the earning while learning problem. Next, in Section \ref{methods}, we demonstrate how to adapt the work \cite{elreedy2021novel} to the single leg problem, allowing their method to be applied to airline RM. Following, in Section \ref{experiments}, we present the experimental results comparing the adapted method to RMS in a simulation environment. We finish our work by discussing future work and conclusions in Section \ref{conclusions}.

\section{Background in airline RM} \label{background}
We consider the single-leg problem, which the airline operates one leg from point A to point B, and a new flight depart at every time step. We assume that the flight's capacity is unconstrained and the time is finite with horizon $H$. The system is allowed to choose fares from a fenceless fare structure with $n$ price points, where $f \in \mathcal{F} = \{f_0, \dots, f_{n-1} \}$ is a price point. The RMS goal is to select prices that maximize the expected revenue $R(f) = f \cdot d(f)$ for each active flight, where $d(f)$ represents the expected number of bookings for price point $f$. A demand model extensively used in the literature \cite{fiig2010optimization,gallego1994optimal} is the negative exponential demand model $d(f;\nu,\phi) = \nu e^{-\phi(f/f_0 - 1)}$, where customers arrive according to a Poisson process with mean $\nu$ and $\phi$ represents the customer price sensitivity. The constant $f_0$ is the base price for which any arriving customer purchases with probability 1.  Since we assume unconstrained capacity and a discrete fare structure, the optimal fare can be trivially obtained with $f_* = \argmax_f R(f)$.

As the RMS does not have access to the true demand behavior parameters $(\nu_*,\phi_*)$, it needs to estimate them from historical bookings. For simplicity, we assume that RMS knows the true arrival rate $\nu_*$, and it needs to estimate only the price sensitivity $\phi_*$. Typically, the parameter estimation relies on statistical techniques such as ordinary least squares or maximum likelihood estimation.

\begin{figure}[t!]
	\centering
	\includegraphics[width=0.6\textwidth]{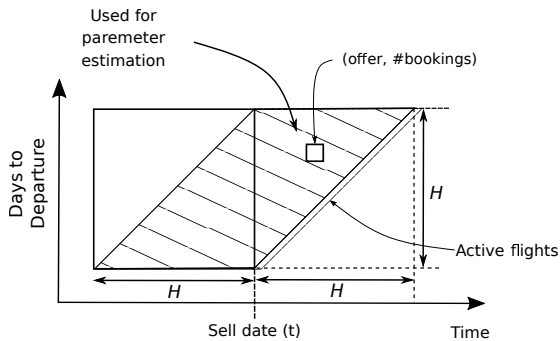}
	\caption{The historical database scheme. Each entry in the database consist of one offer (i.e., the selected fare) and the number of observed bookings for that offer. We keep entries for each day to departure and for each flight. At the end of every day, the booking data collected for that day are immediately added to the historical data and the oldest record is erased.}
	\label{historical-database}
\end{figure}

One specificity of the single-leg problem is related to how the historical bookings are collected. The selling horizon for each flight is divided into $H$ time steps. For each time step, RMS needs to take a pricing decision for each of the $H$ active flights (each departing in a different date). Because the historical database has a fixed size, when new data are appended, the oldest flight data are removed (first-in, first-out). Every time step, the system calibrates the parameters of the demand model from historical bookings and it uses the model to optimize the pricing policy. Given the simplicity of our demand model (i.e., no seasonality, nor day of week, etc.), we assume that the historical database contains the booking data collected for the last $H$ sell dates. The historical database layout is illustrated in \figref{historical-database}.

To compute the price sensitivity parameter $\phi$ and its uncertainty $\sigma$, we represent the historical booking data with the help of two functions. The first one, $o(t, f) \in \mathbb{N}_0$ returns the number of times fare $f$ is offered across all  flights at the sell date $t$. Similarly, the second one, $b(t, f) \in \mathbb{N}_0$ returns the number of bookings made for fare $f$ at the sell date $t$ across all flights at the sell date $t$. Note that the number of customer arrivals is not directly observable (it is not possible to distinguish customers not willing to purchase at price $f$, from no customer arrival at all).

According to \cite{newman2014estimation}, for sell date $t$, we can write the log-likelihood for the negative exponential model as

\begin{equation}
\label{mle}
\mathcal{LL}(\phi) = \sum_{f\in\mathcal{F}}{\sum_{\tau=t-H}^{t-1}  \left[ b(\tau, f) \ln\left(d\left(f\right)\right) - o\left(\tau, f\right) d\left(f\right) \right]}
\end{equation}

then, we can compute the estimated price sensitivity with $\hat{\phi} =  \argmax_\phi \mathcal{LL}(\phi)$.

To compute the estimated error $\sigma(f)$, we can use the inverse of the Fisher information $\sigma^2 \ge 1 / \mathcal{I}(\phi)$, given by

\begin{align}
\mathcal{I}(\phi) &= -\mathbb{E}\left[ \frac{\partial^2}{\partial^2 \phi} \mathcal{LL}(\phi) \bigg| \phi \right] \nonumber \\
&= \sum_{f \in \mathcal{F}}{\sum_{\tau=t-H}^{t-1}  o(\tau, f) d(f) \left(\frac{f}{f_0} - 1\right)^2}
\label{error}
\end{align}

\section{Earning while Learning} \label{related-work}
The common practice in industry consists in estimating the parameters of a demand model, and then assume the unknown parameters are identical to the most recent estimates when optimizing prices is in the roots of EWL problem. It has been demonstrated theoretically that this approach can lead to \textit{incomplete learning}, that is, the true demand model parameter estimates never converge to their true values, independently of how much historical data is collected by the system. To solve incomplete learning, the system must follow a ``semi-myopic'' policy, that accumulates (Fisher) information in an adequate rate, and that do not deviate ``too much'' from the greedy policy \cite{keskin2014dynamic}.

The investigation of the trade-off between earning and learning has generated a stream of studies and ideas \cite{besbes2009dynamic,ferreira2018online,harrison2012bayesian,keskin2017chasing,lobo2003pricing}. Every proposed method shares the two principles of accumulating information in appropriate rates without deviating too much from the greedy policy. However, correctly finding the optimal balance proved to be a tricky task.

One of the most effective and influential methods is called controlled variance pricing (CVP) \cite{den2014simultaneously}. The idea behind the CVP algorithm is to impose a constraint to the greedy pricing policy that the selected prices are not too close from the average of previously selected prices, guarantying sufficient price dispersion. The forbidden range around to the average of past selected prices is called a \textit{taboo interval}. As the amount of the historical data increases, the taboo interval shrinks. In principle, if the amount of historical data grows to infinity, the interval shrinks to zero and the system converges to the optimal price.

More recently, a novel method \cite{elreedy2021novel} combines the revenue maximization and the accumulation of information into a single objective function as

\begin{equation}
\label{from-elreedy}
U(f) = R(f) - \eta \left( \frac{\sigma(f)}{\phi} \right)
\end{equation}

where $\eta$ is an exponentially decreasing trade-off parameter, and $\sigma(f)$ represents the future uncertainty of the demand price sensitivity parameter (how much information $f$ will provide). The system selects the fare that maximizes the objective function  $U(f)$, i.e.\ $f_* = \argmax_f U(f)$. Maximizing objective function is responsible for finding the prices that accumulate more information without deviating too much from the greedy policy. The trade-off parameter regulates the importance the optimization algorithm must give to collecting information of the demand behavior. This method is proven very effective, outperforming the CVP in simulated and real data benchmarks. However, this, like many other studies in EWL, focus on settings where only one product (i.e., one flight) is being optimized. That is, it does not consider the possibility that the data from multiple products can be aggregated to perform model calibration such is the case in the airline industry, where the seats from many flights from the same origin and destination are being sold for different departure dates in parallel. In the next section, we modify this algorithm expressed in \eqref{from-elreedy} to allow the optimization of several parallel flights in the single-leg problem.

\section{Adapting EWL unified objective to single leg problem} \label{methods}
As described in Section \ref{background}, modern RMSs assume that each active flight is independent of the others, optimizing each flight according to $f_* = \argmax_f R(f)$. However, when applying \eqref{from-elreedy} to the single leg problem, it becomes evident that this ``divide to conquer'' strategy is no longer effective because the value that $\sigma$ will assume in the future depends on the aggregated pricing decisions across every active flight. Therefore, we need to adapt \eqref{from-elreedy} to accommodate the possibility of optimizing several parallel flights at once.

To do so, first, we define the pricing policy as a multinomial distribution $\pi = [\pi_0, \pi_1, \dots, \pi_{n-1}]$. Each $\pi_i$ component of this distribution represents the probability of selecting the price point $f_i$ for every active flight for sell date $t$. Then, we re-write the expected revenue $R(f)$ as a function of $\pi$ as

\begin{equation}
R(\pi) = H \sum_{i=0}^{n-1} f_i \pi_i d(f_i)
\end{equation}

and the Fisher information can be written in terms of $\pi$ as

\begin{align}
\label{information-add}
\mathcal{I}(\pi) &= -\mathbb{E}\left[ \frac{\partial^2}{\partial^2 \phi} \mathcal{LL}(\phi) \bigg| \phi, \pi \right] \nonumber \\
&= \sum_{i=0}^{n-1}{ \underbrace{\mathbb{E} \left[ \sum_{\tau=t+H-1}^{t} o(\tau,f_i) \bigg| f_i \sim \pi_i \right]}_{\text{expected historical offers at time}\: t + 1} d(f_i) \left(\frac{f_i}{f_0} - 1\right)^2} \nonumber \\
&= \sum_{i=0}^{n-1}{ \left( \underbrace{\sum_{\tau=t+H-1}^{t-1} o(\tau,f_i)}_{\text{already observed}} + \underbrace{\mathbb{E} \left[ o(t, f_i) \bigg| f_i \sim \pi_i \right]}_{\text{expected future offers}} \right) d(f_i) \left(\frac{f_i}{f_0} - 1\right)^2} \nonumber \\
&= \sum_{i=0}^{n-1}{\left(\sum_{\tau=t-H+1}^{t-1} o(\tau, f_i) + H \pi_i\right) d(f_i) \left(\frac{f_i}{f_0} - 1\right)^2}
\end{align}

In \eqref{information-add}, the term representing the last year of historical offers in \eqref{error}, i.e.\ $\sum_{\tau=t-H}^{t-1}  o(\tau, f)$, is replaced by the historical offers minus the oldest entry (that will be eliminated on the next sell date) plus the expected number of offers that the system is going to make at the current sell date for each price point, i.e.\ $H \pi_i$.

Assuming the lower bound for the uncertainty, $\sigma(\pi) = 1 / \sqrt{\mathcal{I}(\pi)}$, we can re-write the objective function as

\begin{equation}
\label{our-method}
U(\pi) = R(\pi) - \eta \frac{1}{\phi \sqrt{\mathcal{I}(\pi)}}
\end{equation}

We seek a vector $\pi$ that maximizes $U(\pi)$. This can be obtained by solving the non-linear optimization problem

\begin{align*}
    &\argmax_\pi \: U(\pi) \\
    \text{subject to} \\
    & \pi_0, \pi_1, \dots, \pi_{n-1} \ge 0 \\
    & \sum_{i=0}^{n-1} \pi_i = 1
\end{align*}

Once the probability distribution $\pi$ is obtained, the system simply selects the fare for each active flight independently according to $\pi$.

\section{Experiments} \label{experiments}
In this section, we investigate how the performance of our method defined by the optimization metric defined by \eqref{our-method} compares to RMS under the scenario where the price sensitivity parameter is unknown to the system and it must be estimated from historical bookings. To solve the non-linear optimization problem $\argmax_\pi \: U(\pi)$, we find that sequential least squares programming \cite{kraft1988software} is effective and robust. We assume that each flight has infinite capacity (representing, for example, the scenario which the flight's capacity is much larger than the expect number of customer arrivals), and that demand behaves according to the negative exponential model presented in Section \ref{background}. We consider a single fare family with 10 price points $\mathcal{F} = \{\$50, \$70, \$90, \dots, \$230\}$ and that the booking horizon has $H = 22$ days. We perform model calibration at every time step. Recall that, for simplicity, we assume that only the demand price sensitivity parameter is estimated from historical bookings, and the arrival rate parameter $\nu_*$ is known by the system at all time steps.

From now on, we refer to the price sensitivity parameter $\phi$ as the ratio of the lowest fare at which the purchase probability is 50\% (frat5) \cite{belobaba2004algorithms}, which we denoted as $F_5$, and it can be mapped according to $\phi = \ln(2)/(F_5 - 1)$. The frat5 parameter has the convenience of presenting a linear relationship to the price that maximizes $R(f)$ under the negative exponential model assumption. We evaluate both RMS and our method within the interval $F_5^* \in [2.1, 3.8]$, in which the corresponding revenue-maximizing price $f_*$ falls within almost the entire fare structure (except by the most extreme prices, \$50 and \$230, that are never optimal for revenue maximization). From the system's perspective, the range of possible values of the true frat5 parameter is unknown, and its true value could in principle assume any positive value larger than one $F_5^* > 1$. However, to avoid unrealistic estimations for the frat5 parameter, we limit the estimates of this parameter to the range $[1.5, 4.3]$. The real systems often have this and other guardrails. Furthermore, we investigate under a low number of arrivals, $\nu_* = 4 / 22 = 0.18$, where the scarcity of booking data makes it particularly hard to estimate the true price sensitivity from historical bookings, making efficient price experimentation fundamental to the success in the task.

\begin{figure}[t!]
	\centering
	\begin{subfigure}{0.45\textwidth}
		\includegraphics[width=\textwidth]{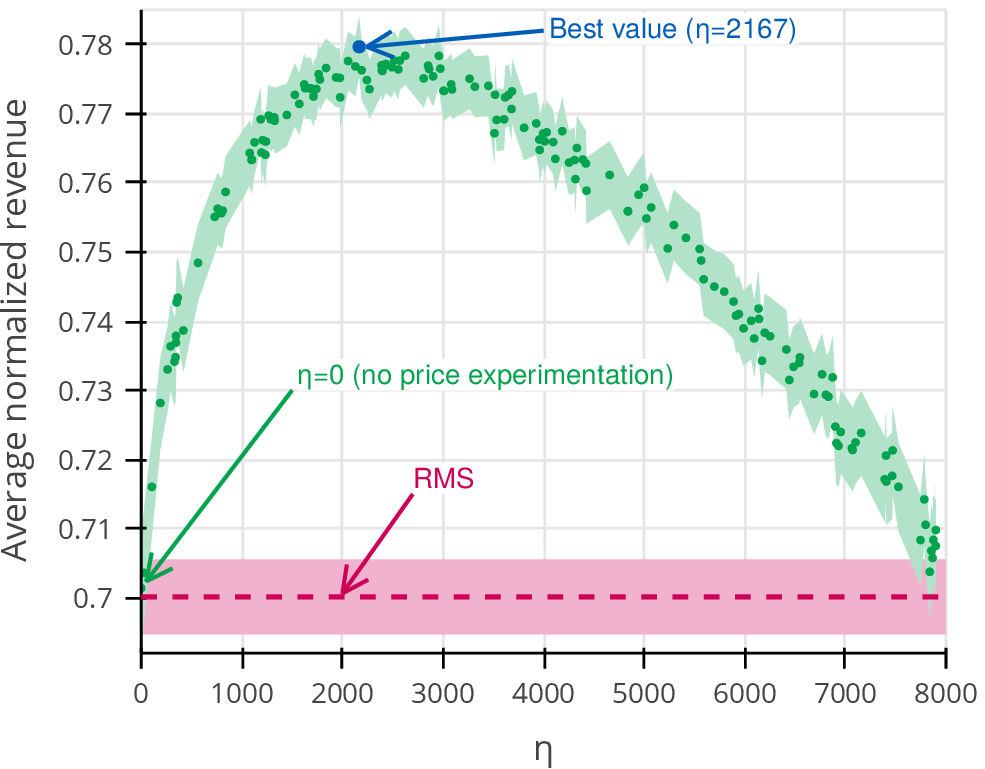}	
	\end{subfigure}
	\begin{subfigure}{0.45\textwidth}
		\includegraphics[width=\textwidth]{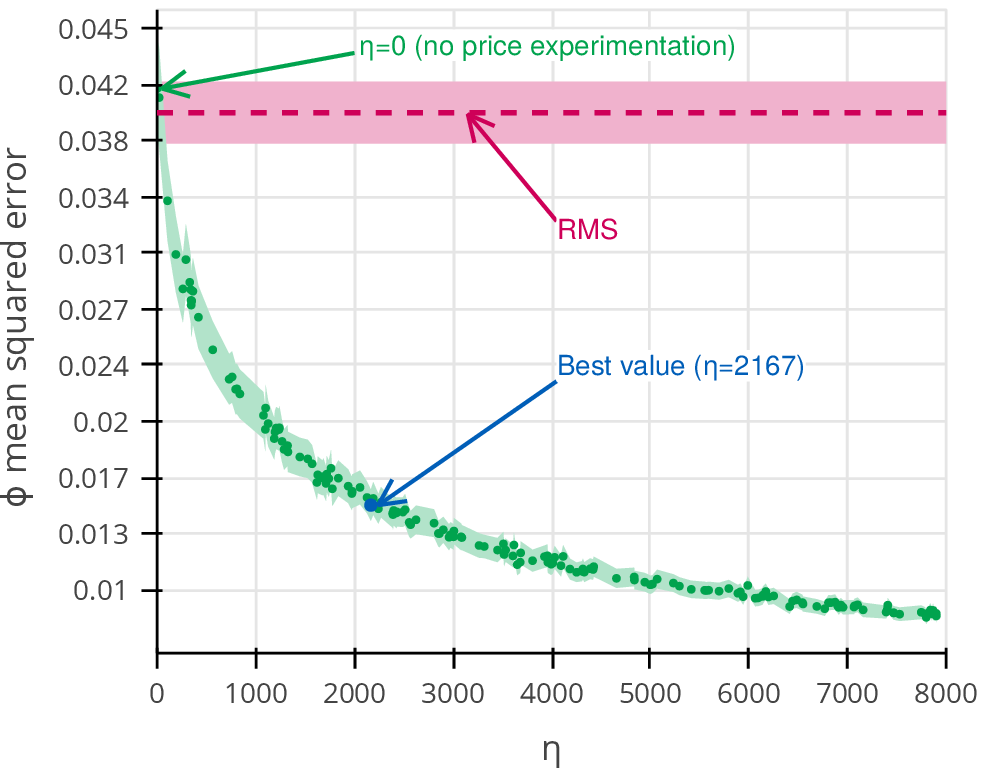}
	\end{subfigure}
	\caption{Optimizing the trade-off parameter $\eta$ (99\% confidence level). \textbf{(Left)} The average normalized revenue normalized with respect to the revenue maximizing policy that knows the true demand price sensitivity at every time step and the random policy that selects fares from a uniformly random distribution. We plot a separate estimation for RMS represented by the dashed horizontal line. \textbf{(Right)} The mean squared error of the price sensitivity estimation $\mathrm{MSE} = \frac{1}{n}\sum_{i=1}^{n} (\hat{\phi} - \phi_*)^2$.}
	\label{eta-opt}
\end{figure}

We now analyze the experimental setup and start discussing the results. We perform three experiments. In the first experiment, we demonstrate how the trade-off parameter $\eta$ influences the revenue output and the quality of the demand model parameter estimation. In the second experiment, we select the best trade-off parameter found in the first experiment, and we analyze how our method and RMS compare with respect to revenue generation and quality of demand price sensitivity estimation for different values of the true frat5 parameter within the evaluation interval. In the third and last experiment, we seek to understand the strategy developed by our method by presenting a detailed comparison between our method and RMS policies and parameter estimation for three distinct values of the true frat5 parameter.

First, we focus on the correct calibration of the trade-off parameter $\eta$. Because we are interested in the system's overall performance within the evaluation interval, for this first experiment, we take the performance metric to be the average collected revenue for the system's pricing policy $\pi$ when the true frat5 parameter is sampled uniformly within the evaluation interval. We sample 160 values for the exploration rate randomly in the interval $\eta \in [0, 8000]$, and for each one of the sampled exploration rates, we sample 2560 values of $F_5^*$ randomly in the evaluation interval and we simulate each episode as described earlier for $20 \cdot H = 440$ time steps. In other words, the performance metric is the average of the revenue obtained by the 2560 episodes of 440 time steps, each having a different true frat5 parameter randomly selected in the evaluation interval. Mathematically, we seek to approximate $\mathbb{E}_{F_5^* \sim [2.1, 3.8]}[R(\pi; F_5^*)]$. In \figref{eta-opt} (left), we show how the system's performance varies with respect to the trade-off parameter. When $\eta = 0$, the system greedily chooses the revenue-maximizing price only, behaving like traditional RMS. As the trade-off parameter increases, the mean squared error (MSE) of the price sensitivity estimation decreases \figref{eta-opt} (right). The better price sensitivity estimation for allows the system to improve its revenue performance. After a point, increasing the trade-off parameter translates to a loss of revenue due to excessive price experimentation. The best revenue performance was obtained at $\eta=2167$, with a normalized expected revenue of $78.0\%$, which represents an absolute improvement of $7.0\%$ when compared to RMS.

\begin{figure}[t!]
	\centering
	\includegraphics[width=\textwidth]{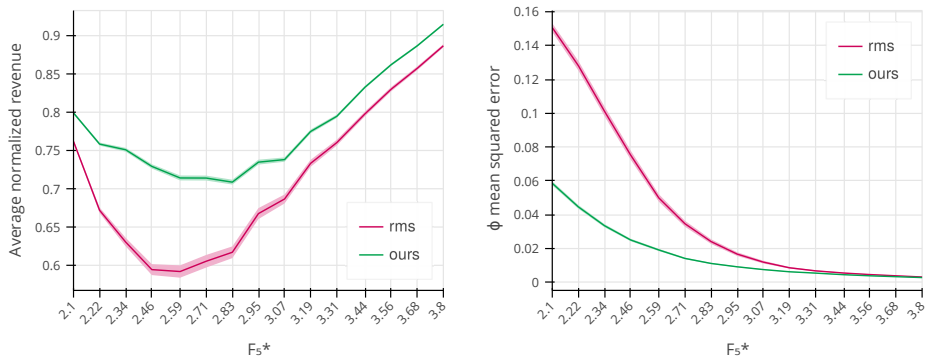}
	\caption{The comparison between the average performance of RMS and our method (99\% confidence level). \textbf{(Left)} The normalized revenue obtained by both methods. \textbf{(Right)} The mean squared error of the price sensitivity estimation.}
	\label{point-performance}
\end{figure}

For the second experiment, we set the true frat5 parameter to a fixed value of the evaluation interval, and for each point we compute the average revenue performance and the price sensitivity MSE of 4000 independent runs when $\eta = 2167$. The comparative results between RMS and our method can be found in \figref{point-performance} (left). We observe that our method performs better than traditional RMS in the entirety of the evaluation interval, with the largest advantage for intermediate values of frat5 $F_5^* \in [2.2, 3.0]$. In practice, this range of the frat5 represents a significant part of the evaluation interval that we are most interested in, because it overlaps with the center of the fare structure, where the RMSs are usually calibrated to operate. In \figref{point-performance} (right), we plot the MSE over the estimation of the price sensitivity. RMS presents larger errors for lower values of the frat5, decreasing as the frat5 increases. Such effect is due to the fact that the true arrival rate parameter $\nu_*$ is known by the system. As the frat5 increases, so does the optimal fare. Further the selected price is from the base fare $f_0$, the higher is the amount of information collected by that price, which can be verified with \eqref{error}. Therefore, the higher is the value of the frat5 parameter, the easier the estimation of the price sensitivity becomes, and naturally, there is less value in price experimentation. Our method, on the other hand, displays much less average MSE on the estimation of the customer's price sensitivity. The effect is more important for lower values of frat5 parameter, where the system's policy approaches the base fare $f_0$, reducing the amount of information obtained by the interaction with the demand. Again, as the frat5 increases, the revenue maximizing price matches the information maximizing price. Consequently, we observe that both methods produce similar results.

\begin{figure}
  \centering
  \begin{subfigure}{0.45\linewidth}
  \includegraphics[width=\linewidth]{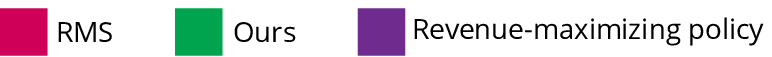}
  \end{subfigure}\par\medskip
  \begin{subfigure}{0.9\linewidth}
  \includegraphics[width=\linewidth]{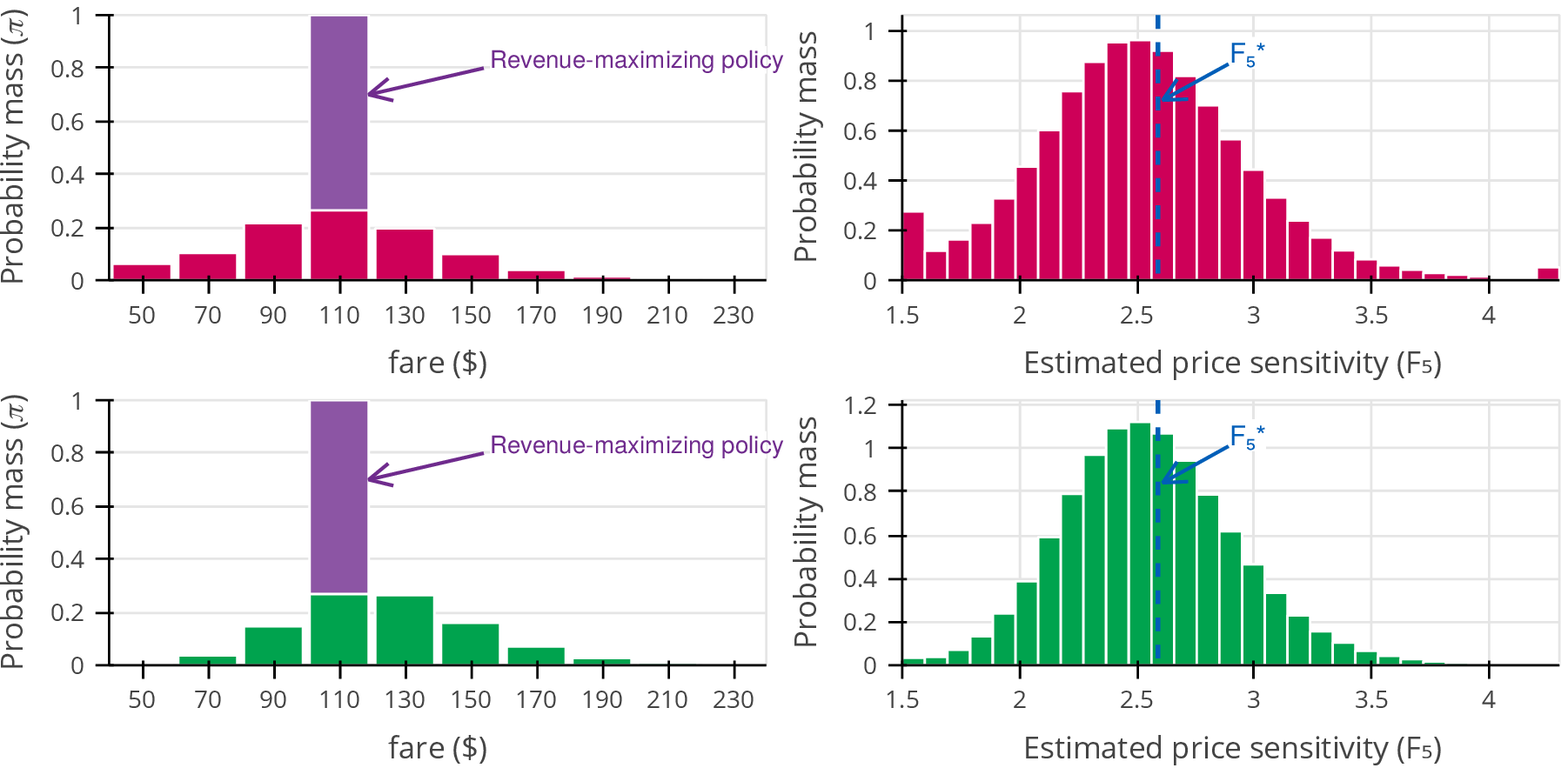}
  \caption{$F_5^* = 2.59$.}
  \end{subfigure}\par\medskip
  \begin{subfigure}{0.9\linewidth}
  \includegraphics[width=\linewidth]{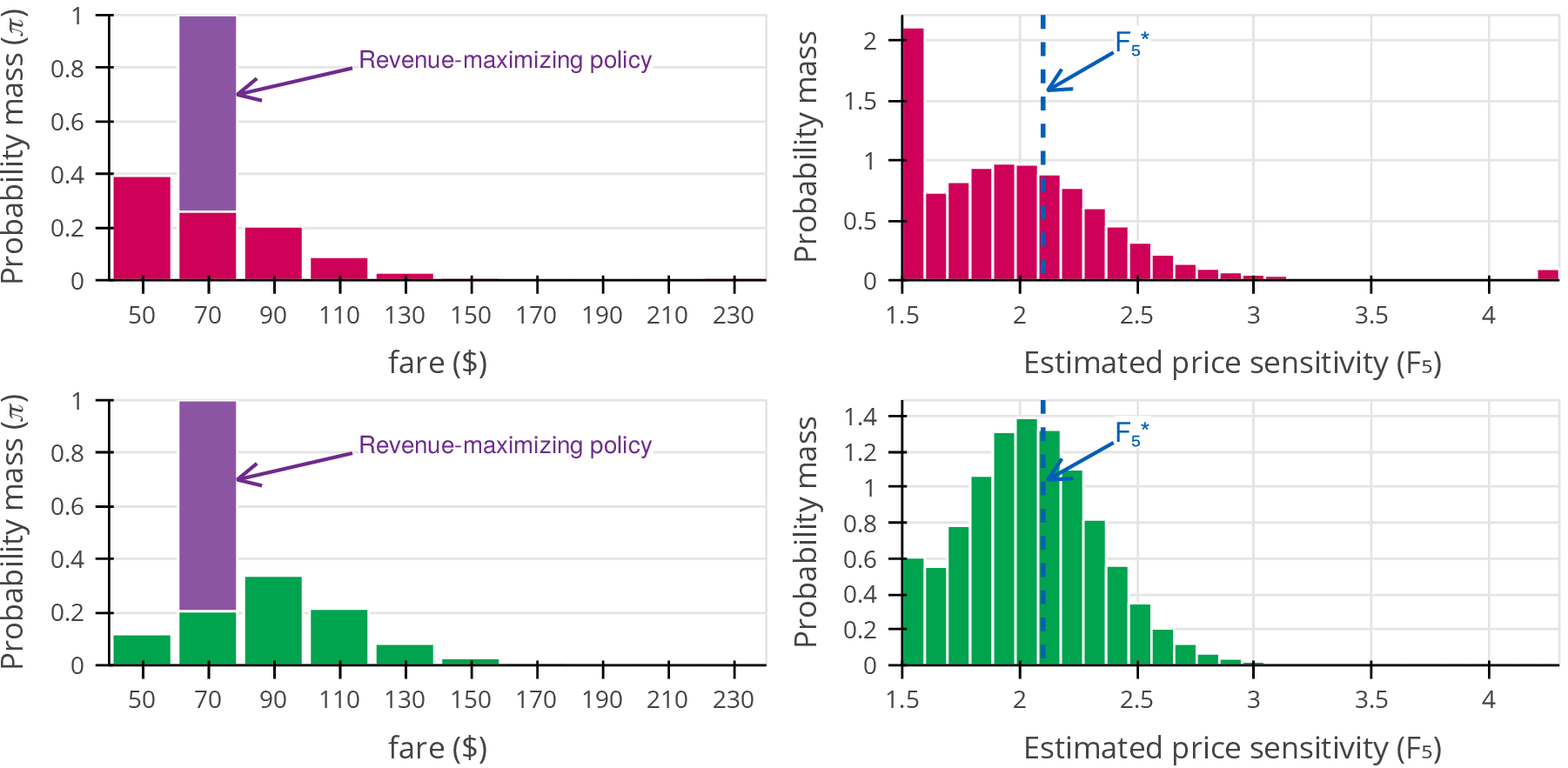}
  \caption{$F_5^* = 2.1$.}
  \end{subfigure}\par\medskip
  \begin{subfigure}{0.9\linewidth}
  \includegraphics[width=\linewidth]{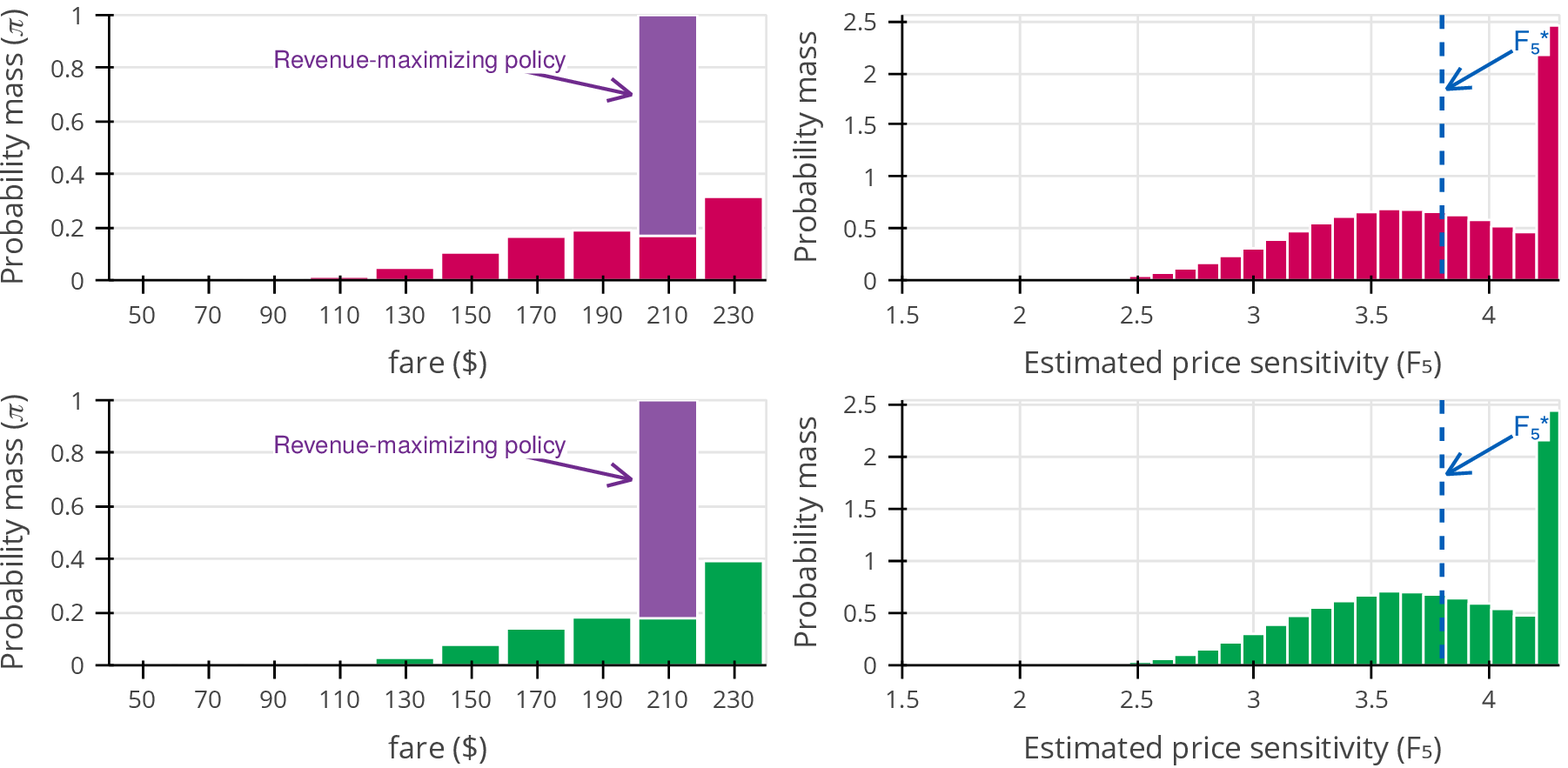}
  \caption{$F_5^* = 3.8$.}
  \end{subfigure}
  \caption{Comparison of the policy and the price sensitivity estimation between RMS an our method. For convenience, we represent the revenue-maximizing policy when the true price sensitivity is known by the system at all times.}
  \label{details}
\end{figure}

Lastly, we perform a detailed analysis of the policy and estimated price sensitivity according to \eqref{mle} for both RMS and our method for three distinct frat5. For the intermediate value of the frat5 parameter $F_5^* = 2.56$ \figref{details}(a), we see on the left figures a comparison between the two policies. In the upper left chart, we can see that RMS tends to price lower fares (\$50, \$70, \$90) rather frequently (38\% of the time), which is a mistake from the perspective of revenue maximization (revenue-maximizing fare $f_* = \$110$), but also a mistake from the perspective of information maximization, because the lower fares provide less information about the demand price sensitivity than the higher fares (see \eqref{error}). Instead, our method shifts pricing towards higher fares that provide more information about the customer's price sensitivity. For the figures on the right, we plot the estimated price sensitivity according to each method. Generally, we see a better estimation of demand price sensitivity for our method (lower right chart), with a narrower distribution around the true value represented by the vertical dashed line. 

For the lower frat5 value, $F_5^* = 2.1$ \figref{details}(b), we see in the upper left chart that RMS tends to price the base fare $f_0 = \$50$ very frequently (40\% of the time), which is neither the revenue-maximizing price (optimal fare $f_* = \$70$), nor appropriate to learn anything about the demand price sensitivity because the base fare provides no information (according to \eqref{error}). On the right upper chart, we observe that the estimated frat5 parameter is often at the minimum accepted value $F_5 = 1.5$, which explains RMS preference for the base fare $f_0$. However, the base fare is near optimal from the revenue-maximizing perspective because of its proximity to the optimal fare. This illustrates the reasons behind RMS surprisingly good revenue performance at low values of the true frat5 parameter that we see in \figref{point-performance}. In other words, RMS is being ``helped'' by the problem settings. Our method, on the other hand, shifts the policy towards higher fares (left bottom chart), which in turn improves the estimation of the frat5 parameter (right bottom chart). As a result, our method provides more revenue on average than the RMS policy while yielding a better estimation of the demand price sensitivity.

For the higher value of frat5 $F_5^* = 3.7$ in \figref{details}(c), we observe a close behavior between RMS and our method. Because the optimal price $f_* = \$190$ provides important information about the customer price sensitivity, there is little gain in experimenting with other prices. In this case, most revenue is lost when the system underestimates the frat5 parameter, selecting lower fares which are both suboptimal for revenue and information maximization.

\section{Conclusions and future work} \label{conclusions}
Correctly estimating the demand price sensitivity is important for RMSs. However, due to little price variation, the RMSs may suffer from incomplete learning of the demand behavior, and as a consequence, produce suboptimal policies.

Inspired by the work developed in \cite{elreedy2021novel}, we present a novel method for jointly optimizing the revenue maximization and model learning under the monopolistic single leg airline RM problem. We show the effectiveness of our method under an unconstrained capacity assumption when estimating a single model parameter (price sensitivity). Our method improves over RMS, which maximizes revenue only, demonstrating that, in some cases, there is value in price experimentation and in the control of model uncertainty.

The value of this research is that it illustrates that airline RMSs may be improved by considering the uncertainty of the demand model parameter estimates during the price optimization. However, because of the unconstrained capacity assumption, this method cannot be applied directly to real-world systems, and further investigation must be done.

Applying our method to real-life airline scenario is challenging for several reasons. First, the unconstrained capacity assumption is not realistic. In practice, airlines have fixed sized aircraft. To maximize revenue under capacity constraints, airlines often use \textit{dynamic programming} \cite{gallego1994optimal}, which is an optimization technique based on the Bellman optimality equation and value functions. Thus, integrating the method here presented with dynamic programming is essential to its applicability in real world problems.

Furthermore, we demonstrate the effectiveness of our method when controlling the error of a single model parameter. However, in real world, demand models may have up to 30 parameters to be estimated and optimized. Obtaining an estimation of the error for each parameter may be not a trivial task. Thus, it will be important to investigate methods for estimating the uncertainty of each model parameter.

Lastly, we calibrated the trade-off parameter $\eta$ by brute forcing values and then selecting the value that works best according to our problem definition. In real world systems, this is very unlikely to be possible. Therefore, we believe that it is important to develop methods for choosing the trade-off parameter that can work under production constraints.

\section*{Acknowledgments}
Thanks to Thomas Fiig, Micheal Wittman, Alexender Papen and Tianshu Yang for the support and useful comments. We would like to also thanks the financial support from Amadeus SAS and the french Industrial Agreements for Training through Research (CIFRE) program.

\bibliography{main}{}
\bibliographystyle{splncs04}

\end{document}